\documentclass[sigconf,nonacm]{acmart}

\setcopyright{none}
\renewcommand\footnotetextcopyrightpermission[1]{}
\usepackage{booktabs}
\usepackage{amssymb} % \checkmark
\usepackage{pifont}  % \ding{55} (cross)
\usepackage{pgfplots}
\pgfplotsset{compat=1.18}
\usepackage{amsmath}
\usepackage{enumitem}
\begin{document}

\title{When Offline Evaluation Misleads: A Diagnostic Protocol for
Reward and Policy Selection in Delayed-Feedback Contextual Bandits}

\author{Sang Su Lee}
\email{psulee@thumbtack.com}
\affiliation{\institution{Thumbtack, Inc.}\city{San Francisco}\state{CA}\country{USA}}
\author{Vineeth Loganathan}
\email{vloganathan@thumbtack.com}
\affiliation{\institution{Thumbtack, Inc.}\city{San Francisco}\state{CA}\country{USA}}
\author{Shishir Dash}
\email{shishirdash@thumbtack.com}
\affiliation{\institution{Thumbtack, Inc.}\city{San Francisco}\state{CA}\country{USA}}
\author{Vijay Raghavan}
\email{vraghavan@thumbtack.com}
\affiliation{\institution{Thumbtack, Inc.}\city{San Francisco}\state{CA}\country{USA}}

\begin{abstract}
Personalizing marketing messages with contextual multi-armed bandits (CMABs) drives real business value,
yet the objective that ultimately matters---a downstream \textbf{conversion}---is observed only weeks
later, too late to drive online learning. Teams therefore train the bandit on a fast \emph{proxy reward},
and separately must judge whether a \emph{contextual} bandit is worth its complexity over sending one best
message. Settling both decisions with the usual offline checks---a batch off-policy estimate, a marginal
arm-discrimination test, a confidence interval---can mislead systematically under delayed feedback. We
give an ordered diagnostic protocol that screens a reward-and-policy candidate on two axes,
\textbf{alignment} (does optimizing the reward move the north-star?) and \textbf{learnability} (can the
bandit identify the reward-optimal policy?), before trusting any reported lift. We \emph{validate} it
where the truth is known---a public off-policy-evaluation benchmark and a controllable synthetic
generator---and \emph{illustrate} it on a deployed large-marketplace push system (where, with five arms
and one split, the evidence is directional rather than powered). Two lessons recur. \textbf{(N1)} A single
offline number can \emph{mis}-rank rewards: a denser reward signal gives the bandit more to learn from, so
rewards that look tied in a static estimate pull apart once learning happens online. \textbf{(N2)} If you
cannot tell in advance which single message is best, a per-user policy partly just \emph{avoids betting on
the wrong one}---that looks like personalization but is really \emph{robustness}, so a ``personalization
premium'' is easily overstated. Our contribution is methodological rather than algorithmic: the ordered protocol, the two lessons it
surfaces, and the end-to-end experience of applying it to a delayed-feedback CMAB.
\end{abstract}


\maketitle
\renewcommand{\thefootnote}{}
\footnotetext{Accepted at the 5th Workshop on End-to-End Customer Journey Optimization (KDD 2026).}
\renewcommand{\thefootnote}{\arabic{footnote}}

\section{Introduction}
Online marketplaces continually choose \emph{which} intervention to send \emph{which} user---a push
notification, in-app message, email, or promotion---to move them toward a valuable action. Picking the
variant per user is naturally a contextual multi-armed bandit (CMAB): the context is the user state, the
arms are the candidate interventions, and the reward should reflect whether the intervention helped. (Our
deployment is a push-notification system, but nothing in the approach is specific to push.) The business
north-star is a delayed downstream \textbf{conversion}---a purchase, booking, subscription, or qualified
lead---that \emph{matures} only over a long horizon. Because it is observed long after the intervention is
sent, it is far too late to drive online learning and cannot train the bandit directly. One common
response---the one we study---is to train on a fast \emph{proxy reward} observed shortly after the send;
modeling the delay itself is another (Section~\ref{sec:related}). Before tuning the bandit at all, then, two upstream questions decide the outcome: which proxy
reward to optimize, and whether a \emph{contextual} bandit is worth its cost over a fixed best message or
a context-free bandit. The framework is domain-agnostic; we describe all signals by their statistical
role rather than their business meaning, both for generality and because the underlying data is
proprietary.

The obvious way to answer both questions is offline and quantitative. Off-policy estimation is the
standard pre-deployment screen for bandit policies~\cite{saito2021obp,voloshin2021ope}: estimate each
candidate's value with one off-policy estimator on a logged batch, test whether the reward varies across
arms, and read a confidence interval. We find this instinct unreliable for delayed-feedback CMABs in three
specific ways, which we first validate where the truth is known (synthetic ground truth and a public
benchmark) and then trace through a real case study.
\begin{itemize}[topsep=2pt, itemsep=3pt, parsep=0pt]
\item \textbf{The interval trap.} With near-uniform logging and $K$ arms, a deterministic policy matches
only ${\approx}1/K$ of the logged traffic, so per-estimate variance dwarfs any plausible lift. A
casually-computed bootstrap interval can make every candidate look significant when none is; the same
data under a correct interval shows the opposite.
\item \textbf{Marginal versus conditional.} A marginal arm discrimination test can declare a reward
``flat across arms'' precisely when the effect a contextual bandit exploits is \emph{conditional} on the
user---present, but invisible to a test that averages over contexts.
\item \textbf{Batch versus online.} A single batch \emph{value} scores one fixed policy on a fixed batch;
it cannot see how an online bandit \emph{learns over time}, and so can rank two rewards as tied when one in
fact induces faster online learning. (The fix is still offline: a step-by-step \emph{replay} of the same
log recovers the learning trajectory the single number hides---Section~\ref{sec:n1}.)
\end{itemize}

Not all three are about delay \emph{per se}: the interval trap and the marginal screen are \emph{general}
off-policy hazards that delayed feedback only aggravates, whereas batch-versus-online (N1) and the
directionality failure are specific to substituting a \emph{fast proxy} for a delayed north-star. What the
delayed-feedback setting adds is that these hazards compound on exactly the two upstream decisions above.

We organize the response as an ordered protocol that screens \emph{alignment} and \emph{learnability}
before trusting any lift (Section~\ref{sec:protocol}), \emph{validate} it where the truth is
known---on synthetic and open-data ground truth (Section~\ref{sec:groundtruth})---and \emph{illustrate} it
on a real deployment (Section~\ref{sec:wild}).
Our contributions are:
\begin{enumerate}
\item \textbf{An integrated, ordered diagnostic protocol} for joint reward-and-policy selection in the
\emph{delayed-feedback} CMAB setting (Section~\ref{sec:protocol}), composing alignment, tool-fit,
learnability, off-policy hygiene, and delay budgeting into a single pre-deployment screen, with an
explicit account of which steps an offline batch estimate can and cannot certify.
\item \textbf{Two empirical findings} we have not seen demonstrated together for this setting:
\emph{(N1)} reward density governs online learning efficiency, so a static batch estimate can rank as
``tied'' two rewards that separate once a bandit learns from them online---demonstrated on ground-truth
synthetic data, reproduced on an open benchmark, and seen directionally on the deployment
(Section~\ref{sec:n1}); and \emph{(N2)} when the
best single arm is unidentifiable from training data, the contextual policy's edge over single arms
\emph{cannot be cleanly attributed} to per-user personalization---a robustness reading is at least as
consistent, which we confirm on a ground-truth generator whose true personalization headroom is zero and
observe on the deployment (Section~\ref{sec:n2}).
\item \textbf{A dual validation and the operational lessons:} the protocol returns the correct verdict on
ground-truth synthetic data and the mechanisms reproduce there, an independent open-source benchmark---its
real logged data and its synthetic generator---independently reproduces all three failure modes (interval trap,
N1, and N2), and the end-to-end run yields a concrete cold-start and metric-maturation timeline
(Sections~\ref{sec:groundtruth}--\ref{sec:discussion}).
\end{enumerate}
We are explicit about provenance. The individual steps are not new: alignment draws on surrogate validity
and directionality, learnability on reward informativeness, and tool-fit on the value-of-personalization
literature (Section~\ref{sec:protocol} cites these). The only genuinely \emph{new} machinery is narrow---%
\emph{replaying} a randomized log to rank \emph{rewards} (not policies) by the learning they induce, and an
\emph{achievable-ceiling} estimate as a fast personalization-headroom screen. Our contribution is to order
these checks into one screen for the delayed-feedback setting, and to surface with it the two findings N1
and N2; we do not claim the individual tests, only their composition and what running it taught us.

\section{Background and Related Work}
\label{sec:related}
\paragraph{Contextual bandits and off-policy evaluation.}
We use disjoint LinUCB~\cite{li2010linucb} as the contextual learner---a per-arm ridge-regression model of
the reward given the context, with an upper-confidence-bound exploration bonus (``optimism under
uncertainty'': try arms whose value is still uncertain, stop once the data settles them)---and evaluate
offline with self-normalized inverse propensity scoring (SNIPS)~\cite{swaminathan2015snips} and a
doubly-robust (DR) estimator~\cite{dudik2011doubly}. SNIPS reweights each logged outcome by the
target policy's probability of the logged action (normalized by the total weight); DR adds an
outcome-model prediction to that importance-weighted correction and is unbiased if \emph{either} part is
correct. Because our log is a near-uniform randomized A/B test, propensities are known by design and both
estimators are unbiased; we prefer them over variants for unknown or heavy-tailed propensities (shrinkage,
sub-Gaussian, double-machine-learning), which target problems this design does not have, and over a pure
outcome model (the direct method), which reintroduces the model dependence we avoid.

\paragraph{Which estimator answers which question.}
\textbf{Doubly-robust is our main estimator}: it answers the standard off-policy question---the value of a \emph{fixed} policy---and carries
the reward comparison, tool-fit (P2), and interval (P4) analyses. The replayer~\cite{li2011replay} is \emph{not}
a substitute for it but a special-purpose complement for the one question a fixed-policy value cannot
answer---how \emph{fast} a reward lets the bandit learn (Section~\ref{sec:n1}, N1). Because it accepts only
the ${\approx}1/K$ of events a deterministic policy matches, it is data-inefficient and high-variance, so
we read only the \emph{shape and ordering} of its learning curves, never their magnitudes. Ironically,
the very ubiquity of one-number off-policy scoring is what tempts practitioners to rank \emph{rewards} by
it---precisely the N1 failure mode.

\paragraph{Known limitations of off-policy evaluation.}
The known limitations of off-policy evaluation---variance, interval
miscalibration, and estimator-dependent conclusions---are catalogued in benchmark studies and
toolkits~\cite{voloshin2021ope, saito2021obp}, and we make no novelty claim on the existence of these
limitations. Our contribution is to show, in one delayed-feedback CMAB workflow, how they combine to flip a
reward/policy decision, and to give an ordered screen that catches them.

\paragraph{Delayed feedback.}
A line of work models the reward \emph{delay} itself---delayed, censored, and aggregated conversions in
bandits~\cite{joulani2013delayed, vernade2017delayed, pikeburke2018delayed}. We instead keep a fast
proxy and treat delay as the reason a proxy is needed, asking which proxy is learnable and budgeting for
the delay operationally.

\paragraph{Choosing and constructing proxy rewards.}
Surrogate-index methods combine short-term proxies to estimate long-term effects~\cite{athey2019surrogate,
prentice1989surrogate}, and proxy-metric selection chooses a metric from past
experiments~\cite{tripuraneni2024proxy}; the experimentation literature defines what makes a good
A/B metric (directionality, sensitivity, predictiveness)~\cite{kohavi2020trustworthy, dmitriev2016seven,
duan2021surrogate}. \emph{Impatient Bandits}~\cite{mcinerney2023impatient} fuses partial and full outcomes
into a predictive reward model. These works \emph{construct} or \emph{learn} a reward; we sit upstream,
screening which candidate reward (primitive or composite) is alignable and learnable in the first place,
and we use reward \emph{informativeness}~\cite{devidze2024informativeness} to explain why density governs
online learning efficiency (N1). The surrogate \emph{paradox}~\cite{vanderweele2013surrogate}---a proxy
correlated with the outcome at the unit level yet anti-aligned across treatments---is the mechanism behind
our directionality check. Reward shaping and multi-objective scalarization study the \emph{combiner} of
several signals~\cite{ng1999shaping, roijers2013morl}; misspecification and Goodhart analyses bound
\emph{over}-optimization~\cite{skalse2022reward, pan2022effects, manheim2018goodhart}. We borrow the
marginal-versus-conditional distinction from this literature rather than claiming it.

\paragraph{Value of personalization.}
The \emph{value of personalization}---the gain of an optimal targeting policy over the best uniform
action, evaluated from a single randomized log---is exactly what \citet{hitsch2024targeting} estimate, and
\citet{shchetkina2024heterogeneity} characterize when treatment-effect heterogeneity is \emph{actionable}
(i.e.\ when that gain is positive). Our achievable-ceiling check (Section~\ref{sec:synthn2}) is a quick
bandit-screen instance of this same quantity. Our setting adds a complementary observation: when the best
single arm cannot be reliably identified from training data, part of the contextual policy's measured
advantage is \emph{robustness} to that unidentifiability rather than realized heterogeneity, so the
premium depends sharply on whether it is measured against a deployable or an oracle single-arm baseline.

\paragraph{Unbiased offline replay.}
\citet{li2011replay} introduced the replayer for unbiased offline evaluation of bandit \emph{policies} on
uniformly-logged data. We reuse the replayer mechanism but for a different purpose---comparing
\emph{rewards} by the online learning trajectory they induce---which is what exposes N1.

\section{Setting and the Diagnostic Protocol}
\label{sec:protocol}
\paragraph{Setting.}
The protocol applies to any \emph{delayed-feedback} CMAB. A user in state $x$ receives one of $K$
near-uniform interventions $a$ (a message variant, in-app action, email, or promotion); a \emph{fast
proxy} reward $r(x,a)$ is observed shortly after, while the business \emph{north-star} $Y(x,a)$---a
conversion or other valuable action---matures only over a long horizon and arrives much later. Offline
evaluation requires a randomized, known-propensity log; we assume the standard case of a uniform-random
arm (propensity $1/K$) with a temporal train/eval split. The reward-and-policy question is: for which $r$
does maximizing $r$ improve $Y$, can a bandit learn that policy, and is a \emph{contextual} bandit worth
it over a fixed arm or a context-free bandit? These are not the only upstream forks---modeling the delay
rather than proxying it~\cite{mcinerney2023impatient}, how to \emph{combine} signals into one
reward~\cite{jeunen2024multiobjective}, and which off-policy estimator to trust~\cite{voloshin2021ope}
also shape the outcome---but they are the two a single offline number most often decides prematurely, and
the two whose wrong answer is costliest and slowest to undo: a misaligned reward can degrade the north-star
for weeks before anyone notices, and an unwarranted contextual policy pays ongoing modeling and serving
complexity for no real gain. We therefore screen these two first.

\paragraph{The protocol.}
We screen each candidate in a fixed order; a candidate must clear the earlier steps before its lift is
worth reading. Steps P1 and P3 subsume the alignment and learnability conditions a reward must meet; P2,
P4, and P5 address tool choice, estimator hygiene, and timing. The two axes are not arbitrary:
\emph{alignment} instantiates surrogate validity and directionality~\cite{prentice1989surrogate,
vanderweele2013surrogate, tripuraneni2024proxy}, while \emph{learnability} extends reward
\emph{informativeness}~\cite{devidze2024informativeness} from a static property to online learning
efficiency; the two are rarely combined into a single reward-and-policy screen, which is what we do here.
\begin{description}
\item[P1 --- Alignment.] Does optimizing $r$ move $Y$? Beyond unit-level correlation (Spearman
$\rho(r,Y)$ on active users), require \emph{arm-level directionality}: the rank correlation between
per-arm mean $r$ and per-arm mean $Y$ must be positive, or the policy is steered toward the arms that are
\emph{worst} on $Y$ (the surrogate paradox~\cite{vanderweele2013surrogate}). Re-check periodically for
drift.
\item[P2 --- Tool-fit.] Is a contextual bandit warranted? Tier the candidates: a fixed best arm, a
context-free bandit, and a contextual bandit. If a context-free policy already captures the gain, the
contextual machinery is not justified; if only the contextual policy beats random, ask \emph{why}
(Section~\ref{sec:n2}).
\item[P3 --- Learnability.] Can the bandit identify the $r$-optimal policy, and how \emph{fast}? Arm
discrimination (a Kruskal--Wallis screen on $r$ by arm) is an identifiability condition, but it is
\emph{marginal} and easily mis-reads conditional structure. Reward \emph{density}/coverage---how often a
usable reward signal is actually observed, i.e.\ the fraction of events that carry a non-trivial
signal---governs learning efficiency, and---crucially---this is read from a step-by-step \emph{replay} of
the log (still offline, but simulating how the bandit learns online), not from a single batch estimate
(Section~\ref{sec:n1}).
\item[P4 --- Off-policy hygiene.] Compute intervals correctly (bootstrap \emph{percentile}, not a normal
interval on the bootstrap mean), reduce variance with a doubly-robust estimator, and check the
overlap/variance ceiling: with $1/K$ effective matching, a small lift may be \emph{unidentifiable}
offline regardless of the reward.
\item[P5 --- Delay budgeting.] Translate the reward window and the north-star maturation into a
time-to-significance: a $w$-day reward imposes a one-time cold-start blind period, and the north-star's
slow maturation adds a fixed lag, both of which gate the earliest trustworthy readout
(Section~\ref{sec:discussion}).
\end{description}

\paragraph{What a pass means (and what it does not).}
The two axes are \emph{necessary-condition screens}, not pass/fail certifications: they reliably flag
\emph{failure} but only green-light a candidate for a live test. \emph{Alignment} has a clear failure
signal---a \emph{negative} arm-level direction (the surrogate paradox)---and a clear non-failure---a
positive one; in between, with few arms the correlation is often unresolved (our deployment's $\rho=-0.80$
has a wide bootstrap CI, Section~\ref{sec:direction}), so the honest rule is \emph{stop} on a clearly
negative sign and \emph{flag-and-test} otherwise, not declare alignment ``true''. \emph{Learnability}
fails when no signal is exploitable (flat both marginally \emph{and} conditionally) or when the reward is
too sparse to converge within the P5 budget; a pass means the replay curve rises in time, not that a
particular lift is guaranteed. We deliberately do \emph{not} set numeric thresholds---with a handful of
arms and one log they would be false precision; the protocol's value is catching the failures cheaply,
before a live test certifies the rest.

\paragraph{What the offline batch can and cannot certify.}
A static value estimate---even a doubly-robust one---can certify alignment direction (P1), screen
marginal arm discrimination (P3), and, under correct intervals, reveal the variance ceiling (P4). It
\emph{cannot} observe the learning-efficiency benefit of density (P3) or the cold-start dynamics (P5):
those need online replay and a live test. The protocol is therefore a cheap pre-deployment screen that
narrows the field and is explicit about what only a live experiment can confirm.

\paragraph{The protocol, its mechanisms, and our two lessons.} Table~\ref{tab:protocolmap} ties the three
together. Each protocol step is validated against ground truth by one of the four mechanisms of
Section~\ref{sec:groundtruth} (M1--M4). Two of those mechanisms surface results that are
\emph{counterintuitive}---a static batch number mis-ranks rewards (P3) and a contextual ``win'' need not be
personalization (P2)---and these are the two we carry as headline lessons, \textbf{N1} and \textbf{N2}. The
other two mechanisms are confirmations that behave as expected once named: M1 is a positive control (where a
contextual bandit genuinely should win, the screen says so) and M3 fixes the delay timeline (P5). So
``only N1 and N2'' is by design: every step is exercised, but only two carry a surprise worth flagging.

\begin{table*}[t]
\caption{The protocol at a glance: what each step screens, the failure mode it catches, where it is
validated against ground truth (Section~\ref{sec:groundtruth}; M1--M4 are Mechanisms~1--4), and which steps
yield our two highlighted lessons. The four mechanisms each validate a step; two of them surface the
counterintuitive lessons N1 and N2, while the deployment case study (Section~\ref{sec:wild}) illustrates
all of them in production.}
\label{tab:protocolmap}
\small
\begin{tabular}{lllll}
\toprule
Step & Screens for & Failure mode it catches & Validated against ground truth & Lesson \\
\midrule
P1 Alignment    & does optimizing $r$ move $Y$?          & backwards reward (surrogate paradox) & synth + covertype (\S\ref{sec:dirsynth}); deployment & --- \\
P2 Tool-fit     & contextual vs.\ fixed / context-free   & unwarranted personalization          & M1 (positive control); M4               & \textbf{N2} \\
P3 Learnability & dense enough to learn online           & batch tie hides the faster learner   & M2; OBP synth; covertype (Fig.~\ref{fig:n1curve}) & \textbf{N1} \\
P4 OPE hygiene  & honest intervals, $1/K$ overlap        & the interval trap                    & OBP real \& synth; digits; covertype    & --- \\
P5 Delay budget & time-to-significance                   & cold-start mis-budgeted              & M3                                      & --- \\
\bottomrule
\end{tabular}
\end{table*}

\section{Ground-Truth and Open-Data Validation}
\label{sec:groundtruth}
Before turning to our deployment, we establish where the correct answer is known that the failure modes the protocol targets are real and general---not artifacts of one setting. Two such settings let us check each
verdict against ground truth: a synthetic generator we control, and an independent open-source benchmark.
The case study (Section~\ref{sec:wild}) then shows the same failure modes arising in production.
Each subsection below isolates one \emph{mechanism}---a causal phenomenon the protocol must handle (e.g.,
reward density driving online learning speed), \emph{exhibited} on the generator described next and checked
against its known verdict; ``mechanism'' refers to the phenomenon under test, not to the data-generating
process itself. Table~\ref{tab:mechresults} collects the synthetic-generator results that follow.

\subsection{A controllable synthetic generator}
\label{sec:synthgen}
We draw a context segment $s\in\{1,\dots,K\}$ uniformly and log an arm $a$ uniformly (propensity $1/K$).
The outcome is $Y\sim\mathrm{Bernoulli}(0.40)$ when $a=s$ and $\mathrm{Bernoulli}(0.15)$ otherwise, so the
best arm for a user \emph{is} its segment. By construction the \emph{marginal} arm effect is zero (every
arm is optimal for an equal share of users) while the \emph{conditional} effect is large. Knobs control
reward density (the fraction of events on which a signal is observed) and reward delay $\tau$.
Sizes per mechanism (all $K{=}5$ unless swept, $70/30$ temporal train/eval split): Mechanism~1 draws
$N{=}60$k events; Mechanisms~2--3 draw $N{=}120$k; the N2 check (Section~\ref{sec:synthn2}) runs $100$ seeds
of $N{=}30$k each and sweeps $K\in\{3,5,10,20\}$, with the no-structure regime using per-arm means evenly
spaced in $[0.17,0.23]$. The generator is deliberately built to \emph{contain} each effect, so it tests
whether the protocol \emph{detects} known structure---not whether the structure exists; the safeguard
against circularity is that the same effects reproduce on data and tooling we did not design
(Section~\ref{sec:obp}), and that the verdict is stable across the swept $K$ and seeds. The specific
constants are \emph{illustrative design choices}, not estimates: $0.40$ vs.\ $0.15$ is simply a clear
conditional contrast, $[0.17,0.23]$ a narrow band that gives a real-but-barely-identifiable best arm with
no conditional structure, and $K{=}5$ / a $70/30$ temporal split are fixed defaults (swept where noted).
The qualitative verdicts are what we report; they do not hinge on these exact values, as the $K$-sweeps and
$100$-seed runs confirm.

\subsection{Mechanism 1: the protocol recovers the truth}
On this generator the protocol returns the correct verdict at every step. The marginal arm-discrimination
test is non-significant ($p=0.75$), correctly reporting no \emph{marginal} effect; the fixed best arm and
the context-free bandit sit at random ($\approx+0.7\%$ true lift); and only the contextual policy realizes
the conditional structure ($+100\%$ true lift), which the doubly-robust estimator recovers ($+104\%$
estimated). This confirms both the protocol's tool-fit logic (P2) and the marginal-versus-conditional
reading (Section~\ref{sec:margcond}) against a known answer. This is also the protocol's \emph{positive
control}: where a contextual bandit genuinely should win, the screen returns \textsc{go}---a green-light to
proceed to a live test---(the $+100\%$ contextual gain, recovered as $+104\%$), so its later cautions are
calibrated, not blanket pessimism. The
validation is two-sided---the N2 check likewise reads large ($+72$ to $+143$pp) exactly when
personalization is real (Section~\ref{sec:synthn2}), and ${\approx}0$ when it is not.

\subsection{The alignment check, validated: a surrogate paradox on demand (P1)}
\label{sec:dirsynth}
P1 rests on \emph{arm-level} directionality, and the generator lets us manufacture the exact trap it
targets. We set monotone true per-arm north-star values $q_Y$ (arm $K{-}1$ best) and a proxy
$r=Y+b(a)+\varepsilon$ whose per-arm bias $b(a)$ is \emph{anti-aligned} with $q_Y$---the \emph{worst}
north-star arm gets the largest proxy boost. This is a textbook surrogate paradox: at the \emph{unit} level
the proxy still correlates positively with the north-star (Spearman $\rho{=}{+}0.77$), so a naive
unit-level correlation check passes it---yet the \emph{arm-level} rank correlation the protocol actually
uses is $\rho{=}{-}1.0$ (bootstrap CI $[-1.0,-1.0]$), and a policy that optimizes the proxy lands on the
\emph{worst} arm: $-50\%$ true north-star versus random and $-67\%$ versus the best fixed arm---optimizing
the proxy moves the north-star \emph{backward}. A matched \emph{aligned} control (bias in the same
direction as $q_Y$) has the \emph{same} positive unit correlation ($+0.80$) but arm-level $\rho{=}{+}1.0$
and recovers the best arm, so the check has a clean non-failure as well as a clean failure---exactly the
\emph{stop-on-negative, flag-otherwise} rule of Section~\ref{sec:protocol}. The effect is not a small-$K$
artifact: sweeping $K\in\{3,5,10,20\}$ the arm-level $\rho$ stays strongly negative ($-1.0$ to $-0.84$) and
the proxy-optimal policy stays $50\%$ below random throughout. This reproduces the deployment's
$\rho{\approx}{-}0.8$ finding (Section~\ref{sec:direction}) on ground truth: the alignment check fires
precisely when optimizing the proxy would degrade the north-star, while a unit-level correlation alone
would wave it through.

\begin{figure}[t]
\centering
\begin{tikzpicture}
\begin{axis}[
  width=\columnwidth, height=4.2cm,
  xlabel={class-rebalancing strength $\gamma$}, ylabel={correlation with north-star},
  xmin=0, xmax=4, ymin=-1.05, ymax=1.05,
  legend style={font=\footnotesize, at={(0.02,0.04)}, anchor=south west},
  tick label style={font=\footnotesize}, label style={font=\footnotesize},
  grid=major, grid style={gray!20},
]
\draw[gray, dashed] (axis cs:0,0) -- (axis cs:4,0);
\addplot[thick, mark=*] coordinates {(0,1.0)(1,0.64)(2,0.21)(3,-0.14)(4,-0.82)};
\addlegendentry{arm-level $\rho$ (P1 check)}
\addplot[thick, dashed, mark=o] coordinates {(0,0.745)(1,0.623)(2,0.316)(3,0.050)(4,-0.075)};
\addlegendentry{unit-level correlation}
\end{axis}
\end{tikzpicture}
\caption{Directionality on real public data (UCI covertype). As a routine class-rebalancing knob $\gamma$
up-weights rare classes, the proxy-optimal policy's true accuracy falls monotonically from $+50\%$ over the
best fixed arm ($\gamma{=}0$) to $-83\%$ ($\gamma{=}4$). The protocol's \emph{arm-level} directionality
check crosses zero and turns significantly negative---catching the surrogate paradox---while a
\emph{unit-level} correlation stays positive well past the point where the policy already degrades the
north-star. Only the arm-level test flags it.}
\label{fig:dirsweep}
\end{figure}
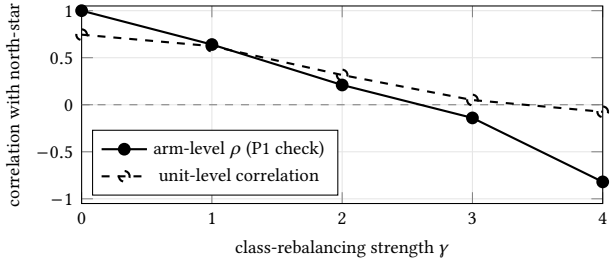

The same paradox arises on \emph{real} public data through a routine mistake, with no hand-injected bias.
On UCI covertype (arms${=}7$ cover-type classes, north-star ${=}$ correct classification, so each arm's
true value is its class prevalence) we take the proxy to be a classifier's confidence and apply one common
``fix'' for imbalance---up-weighting rare classes by $(1/\pi_k)^{\gamma}$ during training. As $\gamma$
grows the proxy increasingly prefers rare, low-value classes and the failure unfolds \emph{monotonically}:
the proxy-optimal policy's true accuracy slides from $+50\%$ over the best fixed arm ($\gamma{=}0$, a
calibrated proxy) to $-27\%$ ($\gamma{=}2$) and $-83\%$ ($\gamma{=}4$). The arm-level directionality the
protocol reads tracks the collapse---$\rho={+}1.0,{+}0.64,{+}0.21,{-}0.14,{-}0.82$---crossing to a
\emph{significantly} negative value (bootstrap CI excluding $0$) by $\gamma{=}3$ and reaching ${-}0.82$ at
$\gamma{=}4$, essentially the deployment's ${\approx}{-}0.8$. Crucially the \emph{unit}-level correlation
stays \emph{positive} well past the point where the policy already hurts the north-star (${+}0.32$ at
$\gamma{=}2$, ${+}0.05$ at $\gamma{=}3$): the unit-level check keeps waving the proxy through while the
arm-level check has already flagged it. A surrogate paradox is thus not a synthetic curiosity---an
everyday class-rebalancing step reproduces it on real data, and only the arm-level alignment test catches
it before launch (Figure~\ref{fig:dirsweep}).

\subsection{Mechanism 2: batch ties, online separates (N1)}
We then build two rewards that are \emph{both} perfectly aligned to the best arm and differ \emph{only} in
density: one observed always, one observed on $10\%$ of events. Trained on the full batch, both recover
the optimal policy, so their batch values are tied ($+0.0$pp gap). Replayed online, the dense reward
accumulates informative signal faster and pulls ahead in the early buckets (a $+50$pp gap before
convergence). Because the ground truth is known and density is the only moving part, this is the
\emph{clean} demonstration of N1; on the deployment (Section~\ref{sec:n1}) the same batch tie becomes a
directional online lead within overlapping intervals, so the synthetic carries the rigor and the real data
the plausibility.

\subsection{Mechanism 3: delay sets the timeline}
Sweeping the reward delay $\tau\in\{0,7,14\}$ days reproduces the cold-start structure we see on the
deployment: the blind period equals $\tau$, the convergence curve shifts right
by $\tau$, and at $\tau=14$ the bandit does not converge within a 60-day window. Operationally,
time-to-significance $\approx \tau + (\text{updates}/\text{rate}) + M$, where $M$ is the north-star
maturation lag---the budgeting rule of P5.

\subsection{Mechanism 4: personalization versus unidentifiability (N2)}
\label{sec:synthn2}
N2 claims that when the best single arm is unidentifiable from training data, the contextual policy's
edge cannot be attributed to per-user personalization. On the deployment this rests on a single split, so
here we validate the \emph{check} itself where the truth is known. We run the generator in two regimes:
a \emph{structured} one (best arm equals the segment, as above), where the true personalization headroom
is large; and a \emph{no-structure} one in which arms carry only small \emph{marginal} differences and the
outcome is independent of context, so the true per-user personalization headroom is exactly zero, yet a
real best arm exists that a finite train split identifies only noisily. In each we read the
achievable-ceiling check of Section~\ref{sec:n2}---the doubly-robust headroom of the contextual oracle
$\arg\max_a \hat q(x,a)$ over the best fixed arm---sweeping the arm count $K\in\{3,5,10,20\}$ (100 seeds,
$N{=}30$k) so the verdict is not a single-$K$ artifact.

The check tracks the truth at every $K$. With no conditional structure the doubly-robust achievable
ceiling stays ${\approx}0$ ($-0.1$pp at $K{=}3$ to $-5.9$pp at $K{=}20$): the contextual oracle does
\emph{not} beat the best fixed arm. With real conditional structure it is large and grows with $K$
($+72$pp at $K{=}3$ to $+143$pp at $K{=}20$). The deployment's near-zero reading (Section~\ref{sec:n2}) is
therefore the \emph{signature of absent personalization}, not a noise or arm-count artifact. Three details
matter. \emph{(i)} Even with zero personalization the contextual policy beats random by $+11.6\%$ (at
$K{=}5$)---but \emph{all} of it is arm selection (it vanishes against the best fixed arm), so a
``personalization premium'' read off the random baseline is spurious by construction. \emph{(ii)} The
train-best arm differs from the eval-best in $31\%$ of seeds ($K{=}5$), confirming the best single arm is
\emph{systematically} unidentifiable, not unlucky in one run. \emph{(iii)} The \emph{direct-method}
(optimistic) ceiling inflates with $K$---up to $+32\%$ at $K{=}20$ despite a true headroom of zero, the
same over-confidence the Direct Method shows in the interval trap (Section~\ref{sec:obp})---which is why
we read the doubly-robust version throughout. This run therefore doubles as a check on the
\emph{estimator}: against the \emph{known} true headroom the doubly-robust ceiling is approximately
unbiased (${\approx}0$ when truly zero, large when truly large) while the direct-method one is not. We
accordingly treat the achievable ceiling as a \emph{conservative screen}---trustworthy for the qualitative
verdict (headroom present or not), not as a precise magnitude.

\begin{table*}[t]
\caption{Mechanism results at a glance, on the controllable synthetic generator (covertype noted where
used); all comparisons are against \emph{known} ground truth. Each row validates a protocol step
(Table~\ref{tab:protocolmap}); the density and ceiling mechanisms are the lessons \textbf{N1} and
\textbf{N2}. ``pp'' is percentage points of true north-star; other figures are relative lift.}
\label{tab:mechresults}
\footnotesize
\begin{tabular}{@{}p{0.18\textwidth} p{0.27\textwidth} p{0.49\textwidth}@{}}
\toprule
Mechanism (step) & Ground-truth setup & Headline result \\
\midrule
Recovers truth (P1,\,P2)        & $K{=}5$, conditional structure              & marginal $p{=}0.75$; contextual $+100\%$ true ($+104\%$ DR); fixed-arm \& context-free ${\approx}$ random \\
Directionality (P1)             & anti-aligned proxy (synth; covertype)       & unit $\rho{=}{+}0.77$ but arm $\rho{=}{-}1.0$; proxy-optimal $-50\%$ vs random, $-67\%$ vs best arm \\
Density $\to$ \textbf{N1} (P3)  & two rewards, same optimum, density differs  & batch gap $+0.0$pp; online gap $+50$pp (dense vs $10\%$-density) \\
Delay (P5)                      & sweep $\tau\in\{0,7,14\}$ days               & blind period $={\tau}$; convergence shifts right by $\tau$; none by $60$d at $\tau{=}14$ \\
Ceiling $\to$ \textbf{N2} (P2)  & sweep $K{\in}\{3,5,10,20\}$, structure vs none & DR ceiling ${\approx}0$ (no structure) vs $+72{\to}+143$pp (structure); DM falsely inflates to $+32\%$ \\
\bottomrule
\end{tabular}
\end{table*}

\subsection{External validity: two independent open sources}
\label{sec:obp}
All three failure modes reproduce on the Open Bandit Pipeline~\cite{saito2021obp}---community tooling and data we
did not write---so none is an artifact of our system or of LinUCB. Table~\ref{tab:datasets} summarizes
every dataset behind this validation: our controllable generator (Section~\ref{sec:synthgen}) plus the
open sources used here.

\begin{table*}[t]
\caption{Datasets behind the ground-truth and open-data validation (Section~\ref{sec:groundtruth}). The
synthetic generator is ours and built to \emph{contain} each effect; the rest are community data and
tooling we did not build. ``clf$\to$bandit'' is the standard supervised-to-bandit conversion (classes as
arms, uniform logging, reward $\mathbb{1}[a{=}y]$), so the full labels give the true policy value.}
\label{tab:datasets}
\small
\begin{tabular}{llccll}
\toprule
Source & Type & $K$ & Size $n$ & Ground truth & What it checks \\
\midrule
Synthetic generator (ours)            & synthetic         & 5 (3--20)            & 30--120k    & by construction      & Mechanisms 1--4 \\
OBP \textsf{SyntheticBanditDataset}   & open, synthetic   & 5                    & 0.5--50k    & known policy value   & interval trap, N1 \\
Open Bandit Dataset (\textsf{ZOZOTOWN}) & open, real logs & ${\approx}80$ (top-3) & ${\approx}10$k & on-policy TS value & interval trap \\
\textsf{sklearn-digits}               & open, clf$\to$bandit & 10                & 1{,}797     & label accuracy       & interval trap \\
\textsf{UCI-covertype}                & open, clf$\to$bandit & 7                 & 15k / 60k   & label accuracy       & interval trap, N1 \\
\bottomrule
\end{tabular}
\end{table*}

\paragraph{The interval trap, on synthetic \emph{and} real logged data.} On OBP's standard
\textsf{SyntheticBanditDataset} ($K{=}5$, context dimension $5$, logistic rewards---OBP's standard
configuration, not a setting we chose; we vary $K$ in the ceiling and directionality sweeps and use larger
$K$ on real data ($\textsf{ZOZOTOWN}{\approx}80$, covertype~$7$); sample sizes $n$ swept
from $500$ to $50$k) we run the full estimator spectrum it exposes
(Replay, DM, IPW, self-normalized IPW, DR, Switch-DR, DR-with-shrinkage, MRDR) against the known value:
the Direct Method returns the \emph{tightest} interval (width ${\approx}0.017$) yet the \emph{largest}
error (${\approx}0.10$) and, unlike the IPW family, its error does \emph{not} shrink as $n$ grows. The same
holds on the benchmark's \emph{real logged data}---the public example sample of the Open Bandit Dataset:
real click logs from \textsf{ZOZOTOWN}, a large fashion e-commerce platform, where a top-$3$
recommendation module over ${\approx}80$ items was served under a uniform-Random and a Bernoulli--TS
policy (${\approx}10$k impressions in the sample; reward $=$ click; ground-truth target $=$ the on-policy
value of the Bernoulli--TS policy, $0.00420$):
DM's $95\%$ interval $[0.00438,0.00474]$ is ${\approx}20\times$ tighter than the IPW family's
($\approx\!0.008$) yet \emph{excludes} the truth, while IPW/SNIPW/DR are wide but cover it (and the
shrinkage estimators collapse toward DM's tight-but-wrong interval). ``A tight interval can be confidently
wrong'' is therefore not an artifact of our pipeline; it holds on independent real open data. We do not
claim DM is intrinsically broken---its error reflects outcome-model misspecification under limited
overlap, and a richer model can shrink it. The point is narrower and robust: a tight-looking interval can
sit around a confidently biased estimate, so an interval must be read against its estimator's
assumptions, not at face value---which is exactly why the protocol pairs it with a variance-honest
estimator and the $1/K$ overlap check (P4).

\paragraph{N1 reproduces (density governs online learning).} Replaying disjoint LinUCB through OBP
synthetic logs ($24$k rounds, $K{=}5$) with two rewards that share the \emph{same} optimal policy but differ only in observation
density, the dense reward accumulates ${\approx}10\times$ more updates and climbs toward the oracle (true
value $0.737$ within a $0.641{\to}0.768$ random-to-oracle range by $18$k rounds), while the $10\%$-density
reward stalls just above random ($0.690$)---the batch-tie / online-separation mechanism of
Section~\ref{sec:n1}, on independent tooling.

\paragraph{N2 reproduces (the achievable ceiling tracks true structure).} Applying the achievable ceiling
check of Section~\ref{sec:n2} to OBP synthetic ($20$ seeds of $N{=}30$k, $K{=}5$): when the reward is
\emph{independent of context} (true personalization headroom $=0$) the contextual oracle's headroom over the best fixed action
is ${\approx}0$ ($-5.9$pp---its entire $+15\%$ edge over random is arm selection), whereas with
context-dependent rewards it is clearly positive ($+16$pp). The check reads ${\approx}0$ exactly when
personalization is truly absent, independently confirming the deployment's near-zero reading.

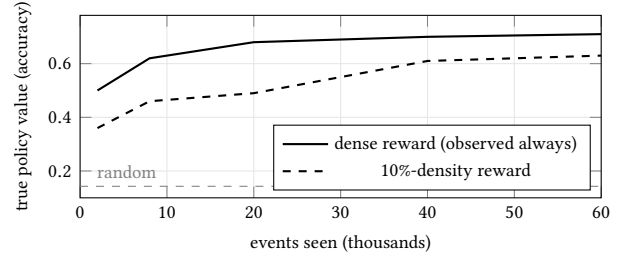
\begin{figure}[t]
\centering
\begin{tikzpicture}
\begin{axis}[
  width=\columnwidth, height=4.0cm,
  xlabel={events seen (thousands)}, ylabel={true policy value (accuracy)},
  xmin=0, xmax=60, ymin=0.1, ymax=0.78,
  legend style={font=\footnotesize, at={(0.98,0.04)}, anchor=south east},
  tick label style={font=\footnotesize}, label style={font=\footnotesize},
  grid=major, grid style={gray!20},
]
\addplot[thick] coordinates {(2,0.50)(8,0.62)(20,0.68)(40,0.70)(60,0.71)};
\addlegendentry{dense reward (observed always)}
\addplot[thick, dashed] coordinates {(2,0.36)(8,0.46)(20,0.49)(40,0.61)(60,0.63)};
\addlegendentry{$10\%$-density reward}
\draw[gray, dashed] (axis cs:0,0.143) -- (axis cs:60,0.143);
\node[font=\footnotesize, gray, anchor=south west] at (axis cs:1,0.143) {random};
\end{axis}
\end{tikzpicture}
\caption{N1 on a public dataset (UCI covertype recast as a bandit, ground truth = label accuracy). Two
rewards point to the \emph{same} optimal policy and differ only in how often they are observed: the
frequently-observed reward learns a better policy at \emph{every} horizon, while the $10\%$-density reward
lags throughout. It is observation \emph{density}, not the reward's target, that governs online learning
speed---the effect a single batch value cannot see.}
\label{fig:n1curve}
\end{figure}

\paragraph{A second independent open source.} These failure modes are not specific to one benchmark either. Under
the standard supervised-to-bandit conversion of \emph{public classification datasets} (classes as arms,
uniform logging, reward $\mathbb{1}[a{=}y]$), the full labels give the \emph{true} policy value (its
accuracy), so we can measure bias directly. On \textsf{sklearn-digits} (handwritten-digit images,
$K{=}10$ classes, $n{=}1{,}797$) and \textsf{UCI-covertype} (forest cover-type from $54$ cartographic
features, $K{=}7$ classes; $n{=}15$k for the interval check, $60$k for the learning curve): the
\emph{interval trap} recurs---the Direct Method returns the
\emph{tightest} interval yet \emph{excludes} the truth (covertype: width $0.009$, ${\approx}6\times$
narrower than IPW/DR, interval $[0.679,0.688]$ vs.\ true value $0.729$), while IPW/DR are wider but cover
it---and \emph{N1} recurs (Figure~\ref{fig:n1curve})---a reward observed on \emph{every} event reaches
accuracy $0.71$ (from a $0.14$ random floor), while the \emph{same} reward observed on only $10\%$ of
events, with ${\approx}10\times$ fewer updates, lags at every horizon. Two of the failure modes, on real
public data we did not generate, with ground truth in hand.

\paragraph{Where each finding is evidenced.} The support is deliberately uneven, and we state it plainly
(Table~\ref{tab:evidence}): the interval trap is the best-supported (real open data and two public
datasets); N1 is strong on ground truth and one public dataset and only directional on the deployment; and
N2's defining \emph{zero-headroom} regime can be \emph{constructed} only in synthetic data, so its
real-world support is the (underpowered) deployment reading alone. Directionality (P1) sits in between: the
surrogate paradox is constructable on demand, so the \emph{check} is validated on synthetic ground truth
and on real public data (a routine class-rebalancing step induces it on covertype, Section~\ref{sec:dirsynth}),
while its only \emph{naturally arising} instance is the deployment. The deployment is illustrative
throughout.

\begin{table*}[t]
\caption{Where each finding is evidenced. \checkmark\ = clear support; (\checkmark) = directional /
underpowered; --- = not exercised. ${}^{\dagger}$N2's zero-headroom case is constructable only in
synthetic data.}
\label{tab:evidence}
\small
\begin{tabular}{lccccc}
\toprule
Finding & our synth & OBP synth & OBP real & public clf & our deploy \\
\midrule
Interval trap        & \checkmark & \checkmark & \checkmark & \checkmark & (\checkmark) \\
N1 (density)         & \checkmark & \checkmark & ---        & \checkmark & (\checkmark) \\
N2 (ceiling)$^{\dagger}$ & \checkmark & \checkmark & ---    & ---        & (\checkmark) \\
Directionality (P1)  & \checkmark & ---        & ---        & \checkmark & (\checkmark) \\
\bottomrule
\end{tabular}
\end{table*}

\section{A Deployment Case Study}
\label{sec:wild}
\paragraph{The deployment.} The case study is a multi-week randomized push-notification A/B test on a large
consumer marketplace: $K{\approx}5$ near-uniform message variants (logging propensity ${\approx}1/K$), a
fast proxy reward, and a $30$-day conversion north-star, evaluated offline on the uniform-random arm with a
temporal train/eval split. With five arms and one split it is \emph{underpowered}, so we treat it as an
\emph{anecdote}---the powered versions of every finding are in Section~\ref{sec:groundtruth}, and the few
numbers below are relative (absolute volumes withheld).

\paragraph{The instructive failure: a backwards reward (P1).}
\label{sec:direction}
The most useful finding needed no lift estimate at all. The deployed reward was a same-day signal that
correlates with conversion at the \emph{user} level, yet at the \emph{arm} level it pointed the wrong way
($\rho{\approx}{-}0.8$): it preferred the arms that converted \emph{worst}---a surrogate
paradox~\cite{vanderweele2013surrogate}---and widening its window flipped the sign. Optimizing it steers
the bandit \emph{against} the north-star, and the directionality check (P1) catches this offline, before
launch. With only five arms this is a directional flag, not a powered test (what is significant is just
that the arms differ on the north-star); but it is exactly the kind of error a lift estimate never sees.

\paragraph{The rest, in brief.}
The other failure modes surfaced here too, but only \emph{consistently} with the ground-truth results
(underpowered on one deployment, not independent evidence): under correct percentile intervals no candidate
reward beat the deployed policy\label{sec:interval}; a marginal arm test called an aligned reward ``flat
across arms'' that a flexible model shows it is not---the conditional effect a context-averaged test washes
out\label{sec:margcond}; a step-by-step replay showed the denser reward learning faster, directionally
(N1)\label{sec:n1}; and only the contextual policy beat random, yet with an achievable-ceiling
personalization headroom of ${\approx}0$---robustness, not realized personalization (N2)\label{sec:n2}. The
lift-independent, defensible change is to \emph{remove the anti-aligned same-day reward} (P1), guarded by a
north-star monitor and an incumbent fallback; among learnable rewards, prefer the denser aligned one.

\paragraph{Prospective check in a subsequent deployment.}
After the analyses above were frozen, the protocol-chosen reward was deployed in a new CMAB test (five
push-message arms, LinUCB, a $10\%$ uniform-random explore floor); the explore slice provides randomized
online data on which we replayed the candidate rewards. The protocol's predictions held: the 1-day reward
would have \emph{confidently} promoted arms that sit mid-to-bottom on the 25-day horizon (a rank reversal,
consistent with the offline anti-alignment finding); the session component left the induced policy unchanged
while roughly tripling the share of users emitting a non-zero learning signal; and the trained policy
achieved significant causal matching over random assignment in the one segment where the arms genuinely
differed ($+43\%$, $z{=}3.11$, against the platform's pre-registered $90\%$ confidence standard). The
deployment's business topline was null; the same data attributes the
null to arm homogeneity in the segment producing the majority of conversions (max$-$min arm spread
consistent with the five-arm noise expectation) rather than to reward selection---no reward choice can
create lift from an undifferentiated arm set. Two limitations follow: the online 25-day ordering is itself
low-powered because the arms are tied, so we treat the replication as structural rather than quantitative;
and the matching result is an exploratory subgroup finding, reported with that label. We did not run the
decisive reward-randomized comparison (protocol-chosen versus naive reward as the randomized unit); it is
the natural next experiment.

\section{Discussion, Recommendations, and Threats}
\label{sec:discussion}
This section steps back from the case study to the general practice. It distills the protocol into a
practitioner checklist, turns it into a deployment plan, and states the threats that bound what we
claim---about the protocol and its ground-truth validation, not only the deployment.
\paragraph{A practitioner checklist.}
For a delayed-feedback CMAB, before tuning the algorithm: (P1) check arm-level \emph{directionality}, not
just unit-level correlation---a same-day proxy can be anti-aligned; (P2) tier fixed-arm, context-free,
and contextual policies to confirm the contextual tool is warranted, and recognize that when the best
single arm is unidentifiable its edge cannot be attributed to personalization alone; (P3) treat marginal arm discrimination as a screen, and read learning
efficiency from \emph{online replay}, not a batch number; (P4) use percentile intervals and a
variance-reduced estimator, and respect the $1/K$ variance ceiling; (P5) budget the cold-start and
maturation lags explicitly.

\paragraph{Turning the protocol into a deployment plan.}
When offline cannot certify a lift over the incumbent---the common case under the $1/K$ variance
ceiling---the protocol still gives directional guidance. Prefer the densest \emph{aligned} reward (N1),
keep the simplest reward that is not statistically separable from it as a low-variance fallback,
warm-start on \emph{recent} data, and \emph{pre-register} the burn-in threshold before any live test.
For timing, budget P5 explicitly: time-to-significance $\approx$ cold-start blind period $+$ enrollment $+$
north-star maturation. The maturation lag is \emph{traffic-independent}, so more traffic shrinks the
minimum detectable effect but not the timeline floor---plan for a smaller effect and a fixed minimum wait
rather than expecting volume to buy speed. (The concrete instantiation for our deployment is in
Section~\ref{sec:wild}.)

\paragraph{Threats to validity.}
A few caveats bound the claims above; we list them roughly in order of how much they limit what we can
conclude.
\begin{itemize}[topsep=2pt, itemsep=3pt, parsep=0pt, leftmargin=1.2em]
\item \textbf{Single real setting.} Our deployment evidence is one marketplace; we mitigate by validating
the protocol and both mechanisms on ground-truth synthetic data and an open benchmark, so the lessons are
mechanism-level rather than setting-specific, but external generalization remains to be shown.
\item \textbf{Offline and counterfactual.} The replay runs on the uniform-random half (a counterfactual,
not the live contextual policy) along a single deterministic path, so its CIs are optimistic and only
orderings and cold-start shape are trustworthy; the burn-in threshold must be pre-registered live, not
chosen post hoc.
\item \textbf{Overlapping intervals.} Under correct (percentile) intervals the deployment's candidate
rewards are wide and mutually \emph{overlapping}, so we never claim one is significantly better than
another; the deployment ranks are directional only.
\item \textbf{Reimplemented baselines.} Comparator policies were reimplemented and only lightly tuned and
consume different signals, so the deployment comparison is not a controlled algorithm benchmark.
\item \textbf{Directionality power.} Arm-level directionality rests on only a handful of arms, so on the
deployment it is directional evidence, not a powered test (the \emph{check} itself is validated on ground
truth, Section~\ref{sec:dirsynth}).
\item \textbf{N2's zero-headroom case is synthetic.} The no-conditional-structure regime that anchors N2
(true personalization headroom $=0$) can only be \emph{constructed}; we exercise it on synthetic
generators (ours and OBP's), since any real dataset carries some structure; we therefore show the
achievable-ceiling check reads ${\approx}0$ when the headroom is truly zero, not that a real system has
zero headroom.
\item \textbf{Delay is budgeted, not modeled.} We treat the maturation lag as a timing budget (P5) and as
the reason a proxy is needed at all; we do not model the delay distribution or the non-stationarity of the
proxy-to-north-star relationship over time. The periodic re-screen of arm-level directionality (P1) is a
coarse guard, not a drift monitor; explicit delay modeling and drift-aware directionality monitoring are
natural extensions.
\item \textbf{Uniform-random logging.} The replay and the achievable-ceiling check both consume the
uniform-random slice. Under a heavily skewed operational logging policy, propensity weights concentrate,
the DR oracle's variance grows, and the ceiling estimate degrades accordingly---so the check is best run
on even a small randomized slice rather than on skewed operational logs.
\item \textbf{No automated thresholds.} The protocol's gates are ordered screens with context-dependent
cutoffs (e.g., positive arm-level rank correlation, a pre-registered burn-in), not calibrated numeric
thresholds; automating threshold selection across settings is future work.
\item \textbf{Data confidentiality.} All reported numbers are relative; absolute volumes are omitted.
\end{itemize}

\section{Conclusion}
For a production delayed-feedback contextual bandit, the upstream choices---which proxy reward, and
whether a contextual bandit is the right tool---are too easily decided by a single offline number that
misleads. We do not claim batch estimates or marginal tests are never useful; we show \emph{when and why}
they mislead here, and each lesson is validated where the truth is known. A confidence interval computed
without regard to the off-policy estimator's variance can be tight yet wrong, so check how any interval
was computed (the interval trap, reproduced on the real Open Bandit Dataset). A batch value can rank two
rewards as tied when one in fact learns faster online, so do not rank rewards on batch lift alone (N1). And
a marginal arm test or a single static estimate cannot see conditional structure or online learning, so
neither is a verdict on its own. Screen alignment and learnability in order, confirm tool-fit, and budget
for delay. We validated
this protocol on a real deployment and on ground truth, and surfaced two findings that a standard offline
workflow hides: reward density drives online learning efficiency, so a static estimate can rank as tied
two rewards that separate once a bandit learns from them---shown cleanly in synthetic ground truth and
directionally on the deployment (N1); and a contextual policy's edge need not be per-user
personalization---when the best single arm is unidentifiable, robustness to that unidentifiability is an
equally consistent reading, which reproduces on a ground-truth generator whose true personalization
headroom is zero (N2).

\appendix
\section{Experimental settings}
\label{app:settings}
Table~\ref{tab:settings} collects every setting behind the results, so the constants in the body can be
read in one place. All synthetic constants are illustrative design choices (Section~\ref{sec:synthgen}),
not estimates; all reported numbers are relative.

\begin{table*}[t]
\caption{Experimental settings for every result. ``temporal $70/30$'' is a time-ordered train/eval split.
Synthetic constants are design choices, not fits.}
\label{tab:settings}
\small
\begin{tabular}{lll}
\toprule
Experiment & Key settings & Outcome / reward \\
\midrule
Synthetic M1 (\S\ref{sec:synthgen}) & $K{=}5$, $N{=}60$k, temporal $70/30$ & $Y{\sim}$Bern$(0.40)$ if $a{=}s$ else Bern$(0.15)$ \\
Synthetic M2 (N1) & $K{=}5$, $N{=}120$k, density $\{100\%,10\%\}$ & same optimum, two densities \\
Synthetic M3 (delay) & $K{=}5$, $N{=}120$k, $\tau\in\{0,7,14\}$d & as M1, reward delayed by $\tau$ \\
Synthetic M4 / N2 (\S\ref{sec:synthn2}) & $K\in\{3,5,10,20\}$, $100$ seeds, $N{=}30$k & structured vs.\ none (means in $[0.17,0.23]$) \\
Directionality, synth (\S\ref{sec:dirsynth}) & $K\in\{3,5,10,20\}$, $N{=}40$--$120$k & $q_Y{=}$linspace$(.10,.30)$; $r{=}Y{+}b(a){+}\varepsilon$, $b$ anti-aligned \\
Directionality, covertype (\S\ref{sec:dirsynth}) & $K{=}7$, $n{=}40$k, rebalance $\gamma\in\{0..4\}$ & $Y{=}\mathbb{1}[a{=}y]$; $r{=}$ classifier confidence \\
OBP synthetic (\S\ref{sec:obp}) & $K{=}5$, dim $5$, $n{=}0.5$--$50$k (OBP default) & logistic rewards; known value \\
OBP real / ZOZOTOWN (\S\ref{sec:obp}) & $K{\approx}80$ (top-3), $n{\approx}10$k & click; on-policy TS value \\
sklearn-digits (\S\ref{sec:obp}) & $K{=}10$, $n{=}1{,}797$ & $Y{=}\mathbb{1}[a{=}y]$; label accuracy \\
UCI-covertype, N1 (\S\ref{sec:obp}) & $K{=}7$, $n{=}15$k / $60$k & $Y{=}\mathbb{1}[a{=}y]$; density $\{100\%,10\%\}$ \\
Deployment (\S\ref{sec:wild}) & $K{\approx}5$, temporal split, one A/B & fast proxy; $30$-day conversion north-star \\
\bottomrule
\end{tabular}
\end{table*}

\bibliographystyle{ACM-Reference-Format}
\bibliography{refs}

\end{document}